\documentclass[letterpaper, 10 pt, conference]{ieeeconf}  % Comment this line out if you need a4paper
\usepackage{afterpage}
\usepackage[T1]{fontenc}

\IEEEoverridecommandlockouts    

\usepackage{graphics} % for pdf, bitmapped graphics files
\usepackage{epsfig} % for postscript graphics files
\usepackage{times} % assumes new font selection scheme installed
\usepackage{amsmath} % assumes amsmath package installed
\usepackage{amssymb}  % assumes amsmath package installed
\usepackage{mathrsfs}
\usepackage{xcolor}
\usepackage{booktabs} 
\usepackage{float}
\usepackage[dvipsnames]{xcolor}

\usepackage{cuted}
\usepackage{caption}
\usepackage{tcolorbox}
\usepackage{hyperref} 

\definecolor{MyBlue}{HTML}{4472C4} % 直接使用十六进制码
\definecolor{MyOrange}{HTML}{ED7D31} % 直接使用十六进制码
\definecolor{MyGreen}{HTML}{70AD47} % 直接使用十六进制码

\definecolor{deepPink}{RGB}{255, 20, 147}

\title{\LARGE \bf
{{\color{MyBlue}Omni}{\color{MyOrange}Dex}{\color{MyGreen}Grasp}: Generalizable Dexterous Grasping \\ via Foundation Model and Force Feedback}
}

\author{
    \textbf{Yi-Lin Wei}\textsuperscript{*},
    \, \textbf{Zhexi Luo}\textsuperscript{*},
    \, \textbf{Yuhao Lin},
    \, \textbf{Mu Lin},
    \, \textbf{Zhizhao Liang},
    \, \textbf{Shuoyu Chen},
    \, \textbf{Wei-Shi Zheng}\textsuperscript{ †} \\ 
    School of Computer Science and Engineering,  Sun Yat-sen University,  China \\
    \href{https://isee-laboratory.github.io/OmniDexGrasp/}{\textcolor{deepPink}{https://isee-laboratory.github.io/OmniDexGrasp/}}
}

\begin{document}
\maketitle

\footnotetext[1]{Equal contribution.}
\footnotetext[2]{Corresponding author.}

\begin{strip}
\vspace{-13ex}
\centering
\includegraphics[width=1.0\textwidth]{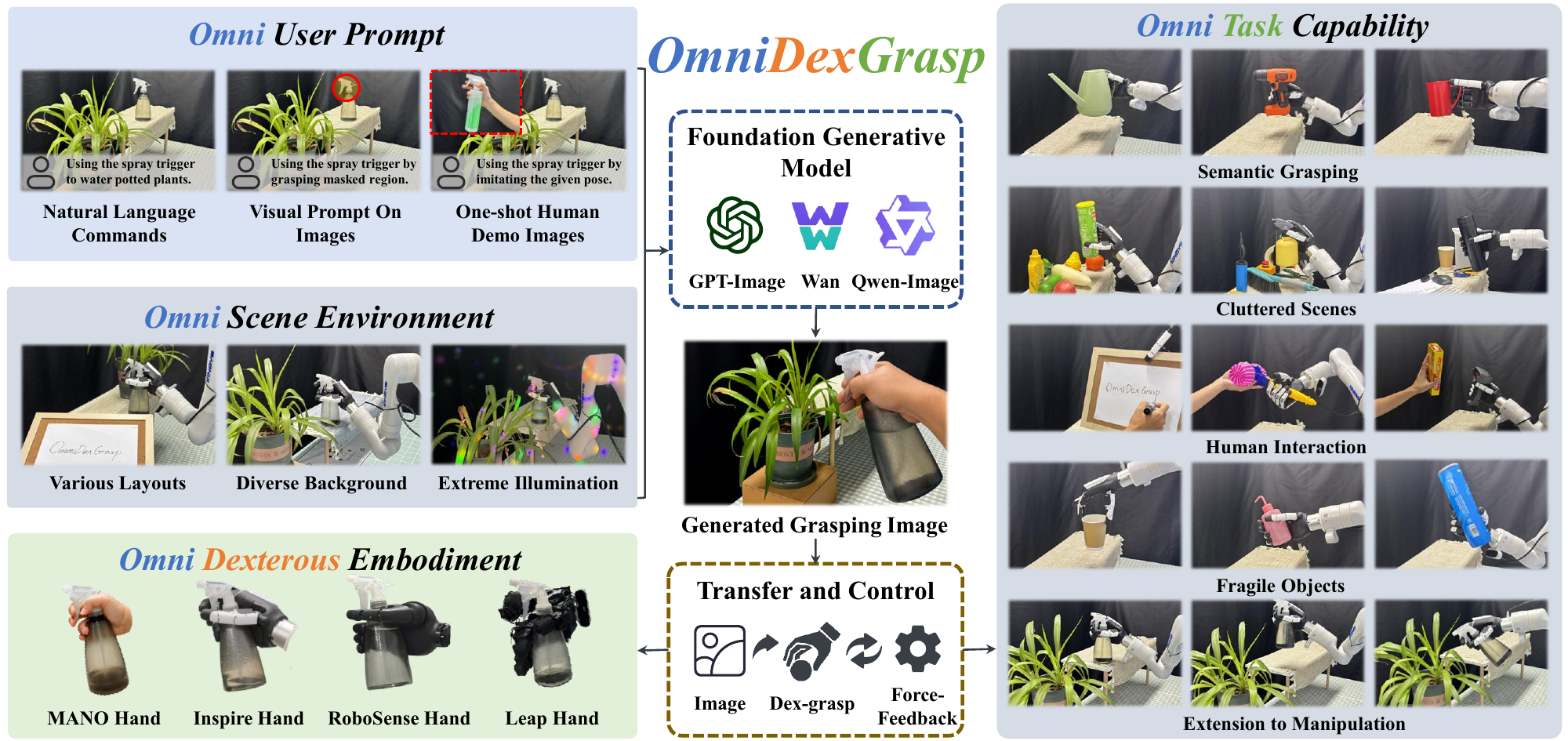}
\normalsize
\captionof{figure}{OmniDexGrasp can achieve generalizable dexterous grasping with omni capabilities in user prompting, dexterous embodiment, scenes, and grasping tasks, by leveraging the foundation model and the propose transfer and grasp strategy.}
\label{fig:setting}
\vspace{-10px}
\end{strip}

%%%%%%%%%%%%%%%%%%%%%%%%%%%%%%%%%%%%%%%%%%%%%%%%%%%%%%%%%%%%%%%%%%%%%%%%%%%%%%%%

\begin{abstract}
\normalsize
Enabling robots to dexterously grasp and manipulate objects based on human commands is a promising direction in robotics. However, existing approaches are challenging to generalize across diverse objects or tasks due to the limited scale of semantic dexterous grasp datasets. Foundation models offer a new way to enhance generalization, yet directly leveraging them to generate feasible robotic actions remains challenging due to the gap between abstract model knowledge and physical robot execution. To address these challenges, we propose OmniDexGrasp, a generalizable framework that achieves omni-capabilities in user prompting, dexterous embodiment, and grasping tasks by combining foundation models with the transfer and control strategies. OmniDexGrasp integrates three key modules: (i) foundation models are used to enhance generalization by generating human grasp images supporting omni-capability of user prompt and task; (ii) a human-image-to-robot-action transfer strategy converts human demonstrations into executable robot actions, enabling omni dexterous embodiment; (iii) force-aware adaptive grasp strategy ensures robust and stable grasp execution. Experiments in simulation and on real robots validate the effectiveness of OmniDexGrasp on diverse user prompts, grasp task and dexterous hands, and further results show its extensibility to dexterous manipulation tasks.

\end{abstract}

%%%%%%%%%%%%%%%%%%%%%%%%%%%%%%%%%%%%%%%%%%%%%%%%%%%%%%%%%%%%%%%%%%%%%%%%%%%%%%%%
\section{INTRODUCTION}
\normalsize
Robotic dexterous grasping stands as a central and highly challenging problem in robotics. The long-term objective is to establish a generalizable and unified framework that enables different robots to understand and execute human commands across a wide range of objects, tasks, and environments \cite{fang2025anydexgrasp, huang2024rekep, wei2024dro, lin2025typetele, huang2023voxposer}. Achieving such capabilities would significantly enhance the practical utility of robots in industrial, domestic, and real-world applications.

With the development of deep learning, data-driven methods show promising performance on diverse objects in dexterous grasp~\cite{2024DGTR, wang2022dexgraspnet, xu2023unidexgrasp}. These approaches based on conditional generative models leverage visual observation and user commands as input to generate grasping actions \cite{wei2024graspasyousay, li2024semgrasp}, trained on datasets containing aligned observation-action pairs \cite{he2025dexvlg, yang2022oakink}. However, the generalization and omni-capabilities of these approaches remain challenging by the scale and diversity of available datasets \cite{wei2025afforddexgrasp, jian2025g-dex}, making it challenging to reliably handle novel object categories, diverse user commands, and dexterous embodiments.

Recent advances in foundation generative models offer a promising pathway to overcome the generalization challenge \cite{wu2025qwen, achiam2023gpt}. Trained on large-scale datasets, these models can learn the knowledge relevant to grasping and manipulation \cite{patel2025video_manipulation, huang2024rekep}. However, their application to robotics remains challenging due to the gap between the foundation models’ high-level knowledge and the low-level physical constraints of robotic execution \cite{tayyab2025foundationrobotics}. As a result, it is challenging for foundation models to directly generate executable robot actions, and the outputs may lead to physically infeasible interactions due to model hallucinations \cite{pan2025omnimanip}.

To address these challenges, we propose OmniDexGrasp, a framework for generalizable dexterous grasping with omni-capabilities in user prompting, grasping tasks, and dexterous embodiment. The motivation of OmniDexGrasp is to exploit the generalization and multi-modal adaptability of foundation generative models by leveraging diverse user prompts to generate human grasp for various grasping tasks without requiring additional robot data. Subsequently, to address the limitation that foundation generative models cannot directly yield robot actions, we introduce a human-image-to-robot-action transfer strategy that covert human grasp images generated by foundation model into executable robot action. Finally, to address the absence of physical constraints in generative models, we develop a force-aware adaptive grasping module that ensures robust and stable execution with appropriate force by controlling the force applied by fingers.

Extensive experiments are conducted in both simulation and real-world environments to validate the effectiveness of the proposed OmniDexGrasp framework. In the real world, we evaluate six dexterous grasping tasks that involve varying objects, user prompts the task objectives. And we also conduct semantic grasping experiments in simulation across objects of different categories. Additionally, we conduct experiments to show that the framework can be extended to dexterous manipulation and flexibly adapted to different dexterous hands.

The main contributions are summarized as follows:

1) We propose the OmniDexGrasp framework, with the insight of leveraging foundation models, to achieve generalizable dexterous grasping with omni-capabilities across tasks, scenes, user prompts, and robot embodiments.

2) We introduce a human-image-to-robot-action transfer strategy that accurately converts visual contents generated by foundation models into executable dexterous robot actions.

3) We present a force-aware adaptive grasp strategy to achieve stable and reliable grasp execution by controlling the force applied by fingers.

\section{Related work}

\subsection{Dexterous Grasp}
Dexterous grasping allows robots to manipulate real-world objects, making it a key topic in robotics research. While early studies \cite{2024DGTR, wang2022dexgraspnet} mainly focused on stability and quality, recent research \cite{li2024semgrasp, wei2024graspasyousay} emphasizes the semantics of grasping, leveraging human language as a condition to guide the task. However, as most of these methods are data-driven, the lack of large-scale and well-semantically-annotated datasets \cite{he2025dexvlg, jian2023affordpose} limits their generalization to diverse real-world objects. Recently, some works\cite{wei2025afforddexgrasp, jian2025g-dex} based on affordance or retrieval show certain levels of cross-category generalization, but their reliance on similar parts shared across categories makes them struggle to extend to truly novel objects with distinct structures in real-world scenarios. To tackle this challenge, we introduce the OmniDexGrasp framework, which leverages pretrained foundation models to transfer their learned grasping and manipulation patterns to generalizable robot execution.

\subsection{Foundation Models for Robotics}
With the rapid progress of computer vision and multi-modal learning, foundation models—such as multi-modal large language models (MLLMs) \cite{achiam2023gpt}, image generation models \cite{wu2025qwen-image}, and video generation models \cite{wan2025wan}, which are trained on diverse multimodal datasets, have shown strong capabilities in capturing underlying patterns and generalizing to guide robotic execution. Recently, some works fine-tunes large language models (LLMs) to build vision-language-action (VLA) models by robot action data \cite{he2025dexvlg, kim2024openvla}, while others \cite{jang2025dreamgen} directly train video generation models on robot video datasets. However, these approaches still depend on robot-specific fine-tuning data, inherently constraining their generalization ability. Alternatively, some works \cite{patel2025video_manipulation, liang2024dreamitate} leverage human videos generated by pretrained foundation models for robotic manipulation with parallel grippers; yet, extending this paradigm to dexterous robotic hands poses greater challenges, especially as model hallucinations resulting in physically implausible interactions. To address these limitations, we propose a novel framework that combines the learning-free utilization of foundation models with a transfer and force-aware strategy, thereby achieving both strong generalization and high-quality dexterous execution.

\section{OmniDexGrasp Framework}
\subsection{Overview}
\textbf{Problem Formulation.} Given the scene single-view RGB-D observation ($\mathcal{O}$ for partial point cloud and $\mathcal{I}^{obs}$ for RGB images) and command of user $\mathcal{C}$ as input, our goal is to generate intention-aligned and high-quality dexterous actions 
$\mathcal{G}^{dex} = (T_{dex}, J_{dex})$, where $T_{dex}$ denotes the 6D pose of arm end-effector, and $J_{dex}$ represents the joint angles of the dexterous hand.

\textbf{Framework Overview.} The OmniDexGrasp framework consists of three main components: (1) human grasp image generation by foundation generative models (Sec. \ref{Grasp Image Generation by Foundation Models}), (2) a transfer strategy that maps human grasp images to executable dexterous robot actions (Sec. \ref{Human-image-to-Robot-action Transfer}), and (3) a force-aware adaptive grasping strategy to achieve stable grasp with appropriate force (Sec. \ref{Force Feedback Control}). Furthermore, we extend our framework to manipulation tasks. (Sec. \ref{Extension to Manipulation}).

\begin{figure*}[t]
    \centering
    \includegraphics[width=\linewidth]{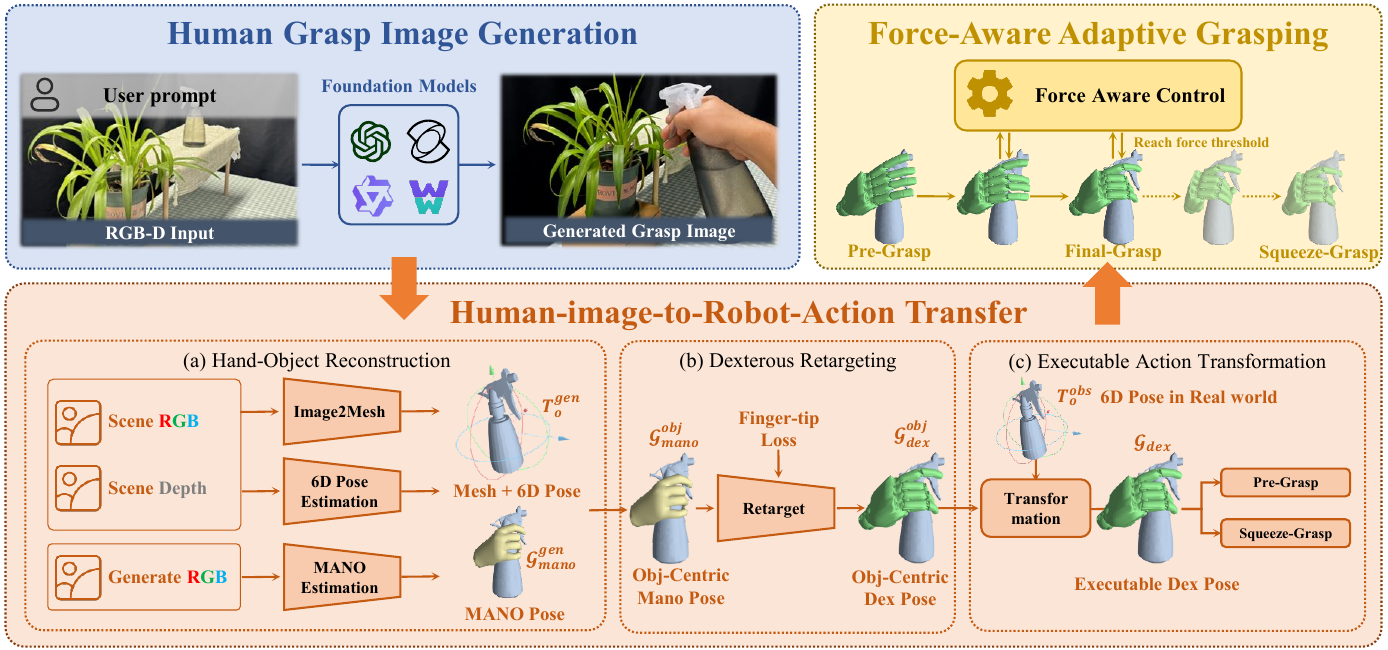}
    \caption{
    Overview of OmniDexGrasp: (1) human grasp image generation via foundation generative models, (2) human-to-robot action transfer mapping grasp images to dexterous robot actions, and (3) force-aware adaptive grasping for stable grasps.
    }
    \label{fig:pipeline}
\vspace{-1em}
\end{figure*}

\subsection{Grasp Image Generation by Foundation Models} 
\label{Grasp Image Generation by Foundation Models}
To achieve generalizable grasping, we first employ foundation generative models \cite{achiam2023gpt} to generate images of human grasping. The advantages of generating human grasps, rather than directly generating robot grasps, are twofold. First, these models are predominantly trained on large-scale datasets rich in human behaviors, which enables them to better exploit generalization capability and prior knowledge of the dynamics and interaction principles. Second, human grasp data serve as an effective intermediate representation that bridges diverse heterogeneous robots and dexterous hands, thereby avoiding the need to train a separate model for each specific robot platform.

Specifically, we feed the RGB observation $\mathcal{I}^{obs}$ and a user prompt into foundation generative models to produce human grasp images $\mathcal{I}^{gen}$. The prompt is omni-capable, supporting diverse forms such as a language instruction, a grasp point or region mask on the observation, and a demonstration image. To further enhance generation quality, we design a prompt template to generate detailed system prompts, including both positive and negative components. An example prompt of language command is shown below:

\begin{tcolorbox}[colback=gray!5,colframe=gray!40!black,title=Prompt Example, fonttitle=\bfseries, boxrule=0.5pt]
% \footnotesize
\scriptsize
\textbf{Positive:} \texttt{Object: \{name\}. Intention: \{intent\}. Based on the input image and grasp intention, generate a image of a human right hand grasping the object. Camera fixed, hand enters from bottom-right, grasps the object, then stays still. Realistic style, uniform lighting, clear details.}

\vspace{0.5em}

\textbf{Negative:} \texttt{Overly saturated colors, overexposed, blurry details, grayish tone, worst quality, low quality, artifacts, ugly, incomplete, extra fingers, poorly rendered hands, deformed, disfigured, malformed limbs, fused fingers}

\end{tcolorbox}

Our framework supports different types of foundation generative models, including both image generation and video generation. For image generation, we directly synthesize grasp images and employ both closed-source models (GPT-Image \cite{achiam2023gpt}) and an open-source model (Qwen-Image \cite{wu2025qwen-image}). For video generation, we extract final frame which is asked to keeping in the grasping action, as the grasp image, utilizing a closed-source model (Kling \cite{kling2024}) and open-source models (WanX \cite{wan2025wan}). We conduct a comparative study (Sec. \ref{sec: different foundation}) to evaluate the performance of different models. Unless otherwise specified, GPT-Image is adopted as the default image generator throughout experiments.

\subsection{Human-image-to-Robot-action Transfer} 
\label{Human-image-to-Robot-action Transfer}
\begin{figure}[t]
    \centering
    \includegraphics[width=\linewidth]{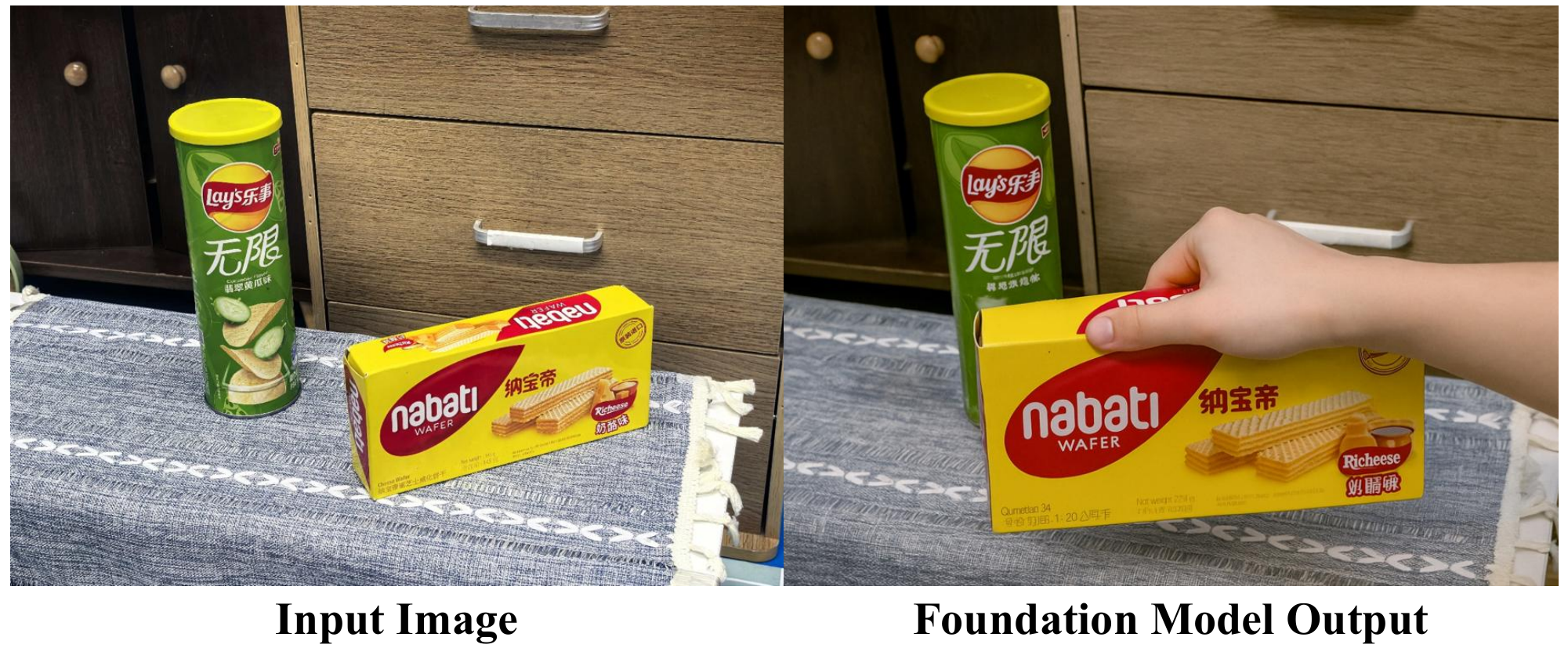}
    \caption{Visualization of the inputs and outputs of foundation generative models, illustrating the potential change of object pose in the generated image.}
    
    \label{fig:vari_pose}
\vspace{-1em}
\end{figure}

Although generating human grasp images is effective, it also introduces the challenge of how to obtain robot dexterous actions, which are both consistent with the images and physically plausible. As shown in Figure \ref{fig:vari_pose}, the object’s generated pose may be altered due to model hallucination. Directly executing the corresponding action often results in failure. To solve this challenge, we propose a human-image-to-robot-action transfer strategy, which inputs human grasp image and outputs executable dexterous action $\mathcal{G}^{dex}$. The pipeline consists of three components: hand-object reconstruction, dexterous retargeting, and executable action transformation.

\textbf{(1) Hand-object Reconstruction.} The aim of this component is to reconstruct the 3D representations of the hand (MANO \cite{romero2022mano} parameters$\mathcal{G}^{obj}_{mano} = (T^{obj}_{mano}, J_{mano})$: 6D pose $T^{obj}_{mano}$ in object coordinate system and articulated pose $J^{mano}$) and the object (object mesh $M_o$ and scale $s_o$) with physically plausible interactions. Specially, for the hand reconstruction, we use HaMeR \cite{pavlakos2024hamer} to obtain articulated pose $J_{mano}$ and wrist 6D pose $T^{gen}_{mano}$ in the camera coordinate system. For object reconstruction, we employ Hyper3D \cite{hyper3d2024}, an image-to-mesh 3D generation model, on the unobstructed original images $\mathcal{I}^{obs}$ to obtain object mesh $M_o$. The object scale $s_o$ is estimated following Any6D \cite{lee2025any6d} by optimization. The object 6D poses $T_{o}^{gen}$ are estimated via a MegaPose \cite{labbe2022megapose} in the generated image camera coordinate system. 

We find that the separate estimation may introduce error in object-hand interaction, which is particularly along the camera’s depth axis. To address this, we optimize the $t_z$ (the translation component of $T^{gen}_{\text{mano}}$ along the depth axis) following EasyHOI \cite{liu2025easyhoi} by aligning hand-object interaction consistency. Finally, we transform the optimized wrist pose into the object coordinate frame by $T^{obj}_{mano} = {T_{o}^{gen}}^{-1} T^{gen}_{mano}$. 

\textbf{(2) Dexterous Retargeting.} The aim of this component is to obtain dexterous hand poses by retargeting the reconstructed MANO human hand $\mathcal{G}^{obj}_{mano}$ to the dexterous hand $\mathcal{G}^{obj}_{dex}$. To achieve this, we employ a two-step optimization that initializes and subsequently refines the dexterous poses. In the first step, the dexterous poses are initialized by copying the wrist's 6-DoF parameters and those of structurally similar joints along the kinematic tree to provide reliable initial values. In the second step, the poses are further optimized in the parameter space by aligning the fingertip positions $p^{dex,ft}_k$ with the corresponding human fingertip positions $p^{mano,ft}_k$. and the optimization objective is formulated as:
\begin{equation}
\label{eq: retar}
\min_{\mathcal{G}_{dex}^{obj} = (T_{dex}^{obj}, J_{dex})} \sum_k \left\| p_k^{dex,ft} - p_k^{mano,ft} \right\|_2^2 .
\end{equation}

\textbf{(3) Executable Action Transformation.} The aim of this component is to obtain executable dexterous poses that are both aligned with real objects and feasible for execution on physical robotic platforms. First, we need align the real object pose with object in generated images, as the generated images do not always correspond precisely to the original scene due to the model hallucination as shown in Figure \ref{fig:vari_pose}. Second, we need to transfer the dexterous pose to robot arm coordinate system to execute action. 

Specially, we estimate the 6D pose $T_o^{obs}$ of object in $\mathcal{I}^{obs}$ by FoundationPose \cite{wen2024foundationpose}. Then, the grasp pose is mapped from the object coordinate system to the real camera coordinate system according to the estimated object pose with transformation $\mathbf{T}_{o \rightarrow c}=T_o^{obs}$. Finally, the grasp is converted from the real camera coordinate system to the robot arm coordinate system via the hand–eye calibration with transformation $\mathbf{T}_{c \rightarrow r}$. 
The complete transformation can be written as:
\begin{equation}
\mathcal{G}_{\text{dex}} 
= \mathbf{T}_{c \rightarrow r} 
\cdot \mathbf{T}_{o \rightarrow c} 
\cdot \mathcal{G}_{dex}^{obj}.
\label{eq: Transformation}
\end{equation}

\subsection{Force-Aware Adaptive Grasping} 
\label{Force Feedback Control}

Although the human-to-robot action transfer strategy generates executable grasp poses, it only provides a target hand configuration. Executing this target in an open-loop manner can easily result in unstable grasps or excessive forces that may damage the object. This limitation arises because the predicted target is not always perfect, due to model hallucinations or the accumulation of errors from earlier stages in the framework.

To address this challenge, we propose a force-aware adaptive grasping strategy based on force-constrained position control. The target grasping force $F_{\text{target}}$ is predicted by a foundation model (e.g., GPT-4o), while the force feedback is measured by the force sensor of each finger (e.g. the strain gauge–based sensor integrated into
the electric cylinder for Inspire Hand). And the target finger configuration is set slightly tighter than the action obtained from the human-to-robot transfer strategy. The grasp is then executed under force-constrained position control, allowing the hand to adaptively adjust its squeezing motion in response to real-time force feedback, thereby ensuring stable and safe dexterous grasping.

Specifically, we first obtain a pre-grasp pose $\mathcal{G}^{pre}_{dex}$ and a squeeze-grasp pose $\mathcal{G}^{squ}_{dex}$ by optimization, where the former provides a collision-free initialization and the latter introduces a more stable grasp contact. The optimization is conducted by adjusting the target contact point $p_k^{\text{mano},ft}$ in Eq.~\ref{eq: retar}. For pre-grasp pose, the contact point is shifted outward by 5 cm along the local surface normal, while inward by 1 cm for the squeeze-grasp pose. During this optimization, the wrist translation and rotation are kept fixed to maintain optimization stability.

Then, the fingers are controlled to move from the pre-grasp pose toward the squeeze-grasp pose under a force-constrained position control strategy. Once the measured contact force reaches the predicted threshold, the current finger position is locked as the new target, preventing further motion to avoid over-squeezing. This strategy enables the hand to adapt to the object’s compliance while ensuring stable and safe grasping. The control output $u(t)$ generated by the PD controller $\mathcal{C}$ is defined as:
\begin{equation}
\begin{aligned}
u(t) &= \mathcal{C}\big(\mathcal{G}^{\text{target}} - \mathcal{G}(t)\big), \\
\mathcal{G}^{\text{target}} &=
\begin{cases}
\mathcal{G}^{squ}_{dex}, & F(t) < F_{\text{target}} \\
\mathcal{G}(t), & F(t) \ge F_{\text{target}}
\end{cases}
\end{aligned}
\end{equation}
where $\mathcal{G}(t)$ and $\mathcal{G}^{\text{target}}$ denote the current and reference joint positions of the finger, $F(t)$ is the force measured by the force sensor.

Finally, the overall grasp is executed in two stages. In the first stage, the arm and dexterous hand move to a pre-grasp pose, where the arm is positioned approximately 10\,cm away from the object along the wrist axis to ensure a safe and collision-free approach. In the second stage, the arm and hand advance together toward the object until the final grasp pose is achieved.

\begin{figure}[t]
    \centering
    \includegraphics[width=\linewidth]{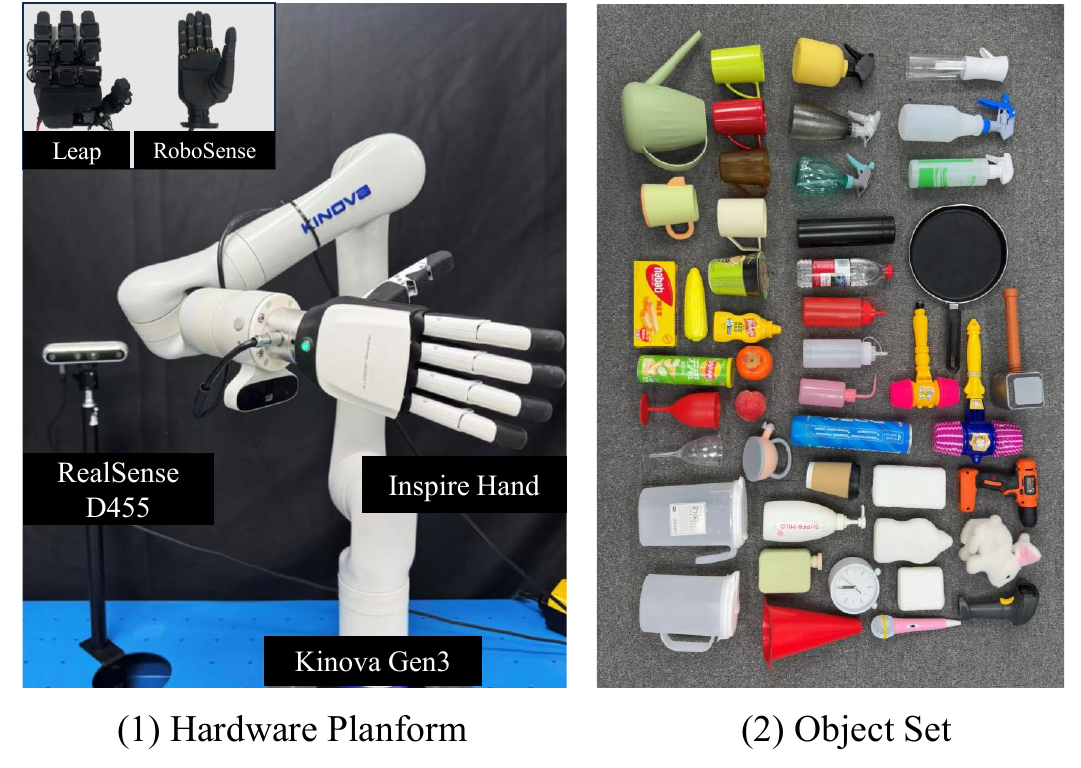}
    \caption{
    Visualization of real world hardware platform and the objects used in experiments.
    }
    \label{fig:setting}
\vspace{-1em}
\end{figure}

\subsection{Extension to Manipulation} 
\label{Extension to Manipulation}
The OmniDexGrasp framework can be naturally extended from dexterous grasping to manipulation tasks, as grasping constitutes a fundamental prerequisite for prehensile manipulation. This extension leverages the generative capabilities of foundation models to reason about object motion post-grasping. Specifically, object motion can be obtained through existing techniques: for instance, by employing an LLM-driven keypoint prediction model \cite{huang2024rekep} to estimate object keypoint movements, or by prompting a generative model to synthesize manipulation videos conditioned on successful grasps and subsequently tracking the object pose in the generated video using FoundationPose. The dexterous manipulation pose is then derived by updating $\mathbf{T}_{o \to c}$ in Eq.~\ref{eq: Transformation} based on the tracked object pose. By coupling these estimated object trajectories with the object-grasp constraint, our framework extends beyond grasping, demonstrating its potential for generalizable embodied intelligence. Qualitative results are illustrated in Figure~\ref{fig:setting} and the supplementary video demo.

\begin{table*}[t]
\centering
\footnotesize
\begin{tabular}{lll cc cc cc}
\toprule
\textbf{Task} & \textbf{Description} & \textbf{Prompt Type} &
\multicolumn{2}{c}{\textbf{ours}} &
\multicolumn{2}{c}{\textbf{w/o Force}} &
\multicolumn{2}{c}{\textbf{w/o Transfer}} \\
\cmidrule(lr){4-5}\cmidrule(lr){6-7}\cmidrule(lr){8-9}
 & & & \textbf{Suc.} & \textbf{Inten.} & \textbf{Suc.} & \textbf{Inten.} & \textbf{Suc.} & \textbf{Inten.} \\
\midrule
Semantic Grasping
  & Grasp guided by Language.
  & Language
  & 83.3
  & 3.82
  & 60.0 
  & 2.90
  & 20.0
  & 0.40 \\

Region / Point Grasping
  & Grasp guided by grasp point or mask.
  & Visual prompts
  & 89.0
  & 3.89
  & 30.0 
  & 1.67
  & 20.0 
  & 0.60 \\

Targeted Grasp in Clutter
  & Grasp user-specified object in clutter.
  & Language 
  & 87.5
  & 4.00
  & 62.5 
  & 3.00
  & 50.0 
  & 0.00 \\

Human-Robot Handover
  & Grasp for handover interaction
  & Language
  & 100.0
  & 4.80
  & 30.0 
  & 1.90
  & 25.0
  & 0.75 \\

One-Shot Grasping
  & Grasp guided by single demonstration
  & Demo image
  & 80.0 
  & 3.80
  & 60.0 
  &  3.11
  &  10.0
  &  0.40\\

Fragile Object Grasping
  & Grasp fragile object safely
  & Language/force
  & 88.0 
  & 4.50
  & 56.0 
  &  2.67
  &  0.0
  & 0.33\\
\bottomrule
\end{tabular}
\caption{The results of the framework on 6 diverse dexterous tasks and the ablation study. "w/o force" means using the poses obtained from
the transfer strategy and executing them with open-loop position control with force feedback. "w/o transfer" means directly using the grasp poses estimated from the generative model.}
\label{tab:tasks_results}
\end{table*}

\begin{figure*}[t]
    \centering
    \includegraphics[width=\linewidth]{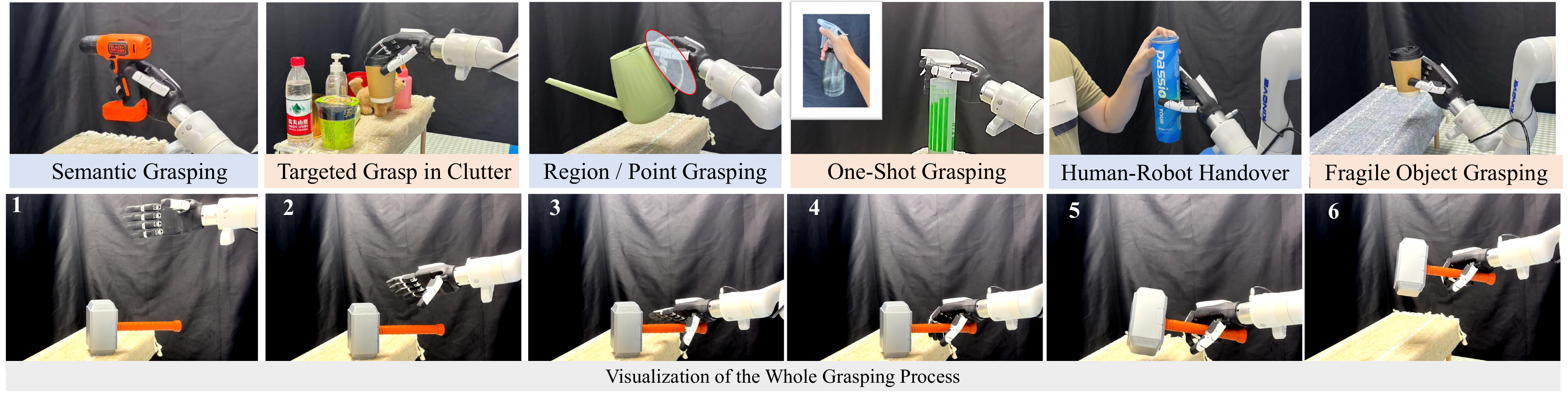}
    \caption{
    Visualization of the six tasks (upper) and the whole grasping motion process (lower).
    }
    \label{fig:vis_exp}
\vspace{-1em}
\end{figure*}

\section{Experiments}

\subsection{Experiment settings}
We conduct diverse experiments on both real world and simulation experiments, we introduce the experiment settings from 1) robot platform, 2) task description, 3) the details of objects used in experiments.

\textbf{Robot platform.} For real world experiments, as shown in \ref{fig:setting}, we conduct all quantitative experiments on a 6-DOF Kinova Gen3 arm, a Realsense D455 camera, and an linkage-driven 6-DOF Inspire FTP Hand. Additionally, we conduct qualitative experiment with fully-actuated 16 DOF Leap Hand \cite{shaw2023leap} and the linkage-driven 8-DOF RoboSense Papert Hand. For simulation experiments, we conduct experiments on the tendon-driven 22-DOF Shadow Hand.

\textbf{Task description.} For real world experiments, we evaluate our framework on six different grasping tasks to assess omni-ability across diverse scenarios and user prompts as shown in Table \ref{tab:tasks_results}. Since no available methods are capable of simultaneously handling such a wide range of tasks, we further conduct extensive and comparison experiments on semantic grasping in \cite{wei2025afforddexgrasp} with representative baselines in both real world (Table \ref{tab:objects_results}) and simulation (Table \ref{tab:simulation_results}).

\textbf{Details of objects.} We details the objects used in our experiments. In our real-world experiments, we used more than 40 objects selected to cover a wide range of shapes, sizes, and functional properties. These objects were grouped into eight categories: handles, spray bottles, liquid containers, packaged food items, cylinders, shafts, fragile objects, and mixed-category objects. In our simulation experiments, we employed over 100 objects from 33 distinct categories \cite{wei2025afforddexgrasp}. The testing objects are divided into three groups: seen, similar, and novel. "Seen" objects belong to categories encountered during model training, "similar" objects come from categories that are related but not identical to the training set, and "novel" objects represent entirely unseen categories. This categorization is designed to enable a more systematic evaluation of training-based grasp models. Importantly, our model has never been exposed to any dexterous robot grasping data during training, highlighting its ability to generalize to unseen objects and categories in simulation.

\subsection{Evaluation Metric} We employ two types of evaluation metrics: grasp success rate and intention-consistency score. For grasp success rate $Suc.$ in real-world experiments, a grasp is considered successful if the robotic hand can lift and hold the object while accurately following the given language command. Furthermore, we employ two GPT-assisted score inspired by \cite{li2024semgrasp}. Specifically, the grasping image and instruction is provided to GPT-4o or human expert, which evaluates two points on a 5-point scale: 1) $Inten.$: the alignment between the executed action and the command; 2) $Stab.$: grasp stability.

\begin{figure}[t]
    \centering
    \includegraphics[width=\linewidth]{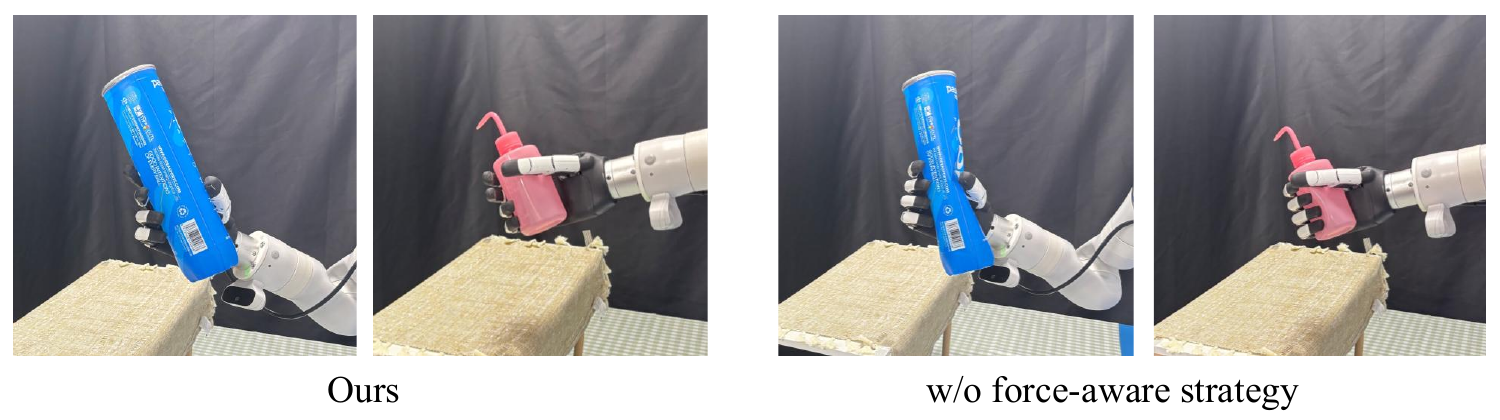}
    \caption{
    The visualization of the ablation of force-aware strategy. Without the force-aware strategy, robotic grasping may damage the objects.
    }
    \label{fig:force_vis}
\vspace{-1em}
\end{figure}

\subsection{Can OmniDexGrasp achieve omni-capabilities across diverse dexterous grasping tasks?}
We evaluate OmniDexGrasp on six representative dexterous grasping tasks (Table~\ref{tab:tasks_results}, Figure \ref{fig:vis_exp}). The framework achieves consistently high grasp success rates (Suc., avg. 87.9\%) and intention-consistency scores (Inten., avg. 4.14), showing strong adaptability to diverse objects, scenes, and prompt modalities. Semantic-dependent tasks (semantic grasping, cluttered grasping, human–robot handover) benefit from foundation model integration, while visual- and demo-prompt tasks (region grasping, one-shot grasping) validate its flexibility on system inputs. Performance on fragile object grasping further highlights its ability to maintain stability while applying appropriate force. Overall, OmniDexGrasp demonstrates omni-ability across diverse tasks within a unified framework.

To the best of our knowledge, since no available dexterous grasping baseline can realize all these omni-capabilities, we do not include direct comparisons with other methods in Table~\ref{tab:tasks_results}. Alternatively, we present extensive comparisons on semantic grasping tasks (Section~\ref{sec: comparison}) through real-robot experiments (Table~\ref{tab:objects_results}) and simulation datasets (Table~\ref{tab:simulation_results}). And we further conduct ablation studies in these six diverse tasks (Table~\ref{tab:tasks_results}) to validate the effectiveness of each component of our method.

\subsection{Are the transfer and force-aware grasping strategies crucial for the framework?}
To validate the effectiveness of the two proposed strategies, we conduct an ablation study, as shown in Table \ref{tab:tasks_results}. The column "w/o Transfer" refers to directly using the grasp poses estimated from the generative model without first transferring them to the object coordinate system and then to the real-world coordinate system. The results show a performance decline because the generative model may produce hallucinations, causing the object pose to change during generation, which can lead to grasp failure. The column "w/o Force" refers to using the poses obtained from the transfer strategy and executing them with open-loop position control. Without force-aware adaptive grasping strategy, the grasp may become unstable if the fingers close too loosely, resulting in an approximate 40\% reduction in success rate in average, or it may damage the object if the fingers close too tightly as shown in Table \ref{fig:force_vis}. Overall, both strategies contribute to enhancing the robustness and stability of robotic grasping across different tasks.

\subsection{Does OmniDexGrasp outperform existing dexterous grasping methods?}
\label{sec: comparison}
We conduct comparison experiments in both real-world and simulation settings to evaluate semantic grasping tasks and perform comprehensive comparisons. All baseline are trained following \cite{wei2025afforddexgrasp}, where the baseline in Table \ref{tab:objects_results} is trained on its complete dataset \cite{wei2025afforddexgrasp} containing over 43,504 samples and 1,536 objects across 33 categories. And the baselines in Table \ref{tab:simulation_results} are trained only on 10 seen categories and evaluated on 11 similar and 12 novel categories for a better evaluation of these generalization. 

\begin{table*}[t]
\centering
\setlength{\tabcolsep}{3.8pt}
\begin{tabular}{l|cc|cc|cc|cc|cc|cc|cc|cc}
\toprule
          & \multicolumn{2}{c|}{Handle} &
            \multicolumn{2}{c|}{Spray Bottle} &
            \multicolumn{2}{c|}{Liquid Container} &
            \multicolumn{2}{c|}{Packaged Food} &
            \multicolumn{2}{c|}{Cylinder} &
            \multicolumn{2}{c|}{Shaft} &
            %  \multicolumn{2}{c}{Easily Deformable Objects}  \multicolumn{2}{c}{Force-Sensitive Objects} \multicolumn{2}{c}{Compliant Objects}
            \multicolumn{2}{c|}{Fragile Objects} &
            \multicolumn{2}{c|}{Mixed Category} \\ 
          & \multicolumn{2}{c|}{
                \raisebox{-0.15\height}{\includegraphics[width=0.054\linewidth]{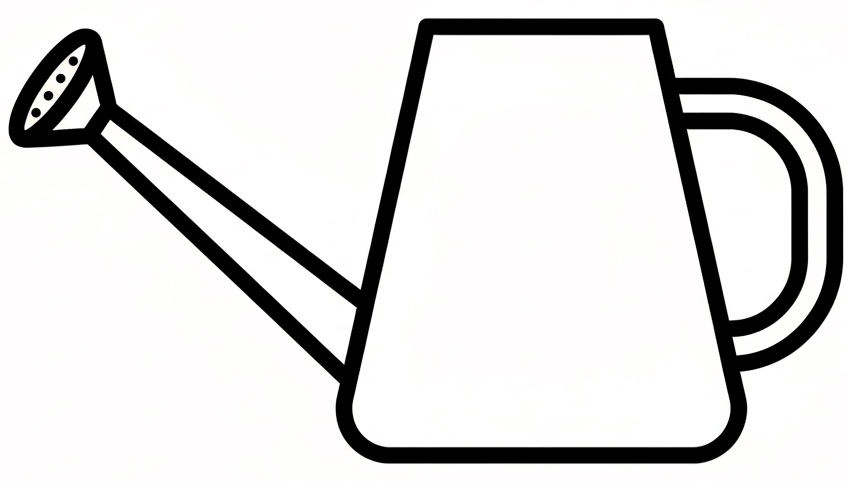}}
                \raisebox{-0.15\height}{\includegraphics[width=0.031\linewidth]{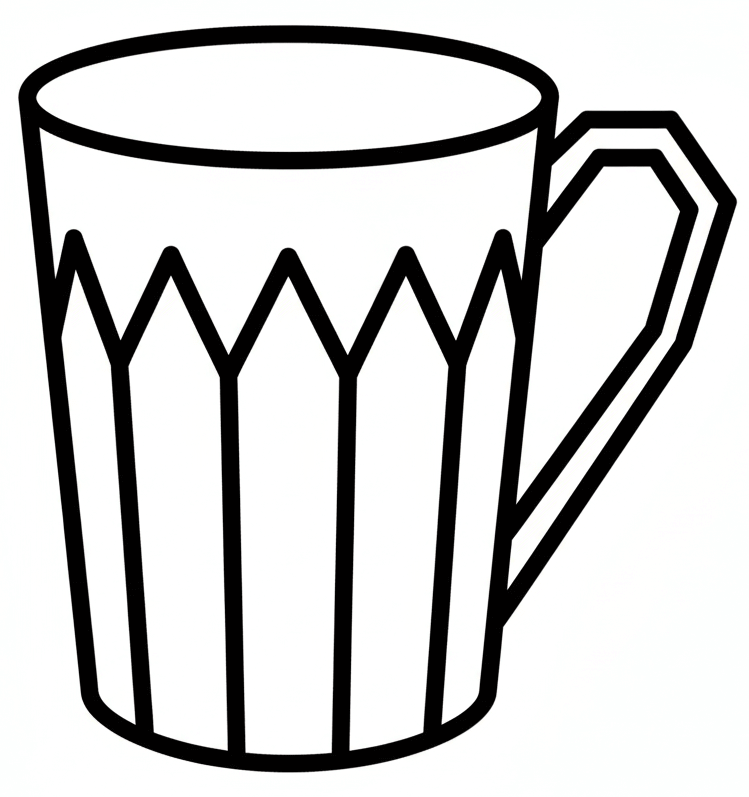}}
                \raisebox{-0.15\height}{\includegraphics[width=0.029\linewidth]{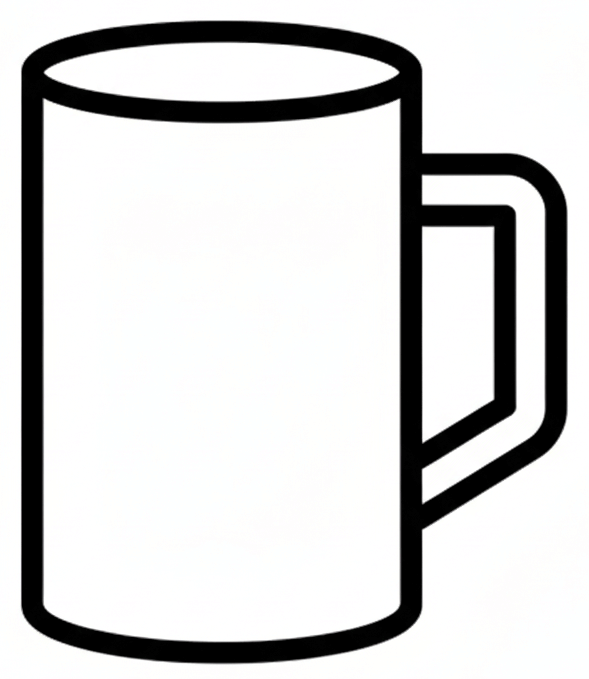}}
            }
          & \multicolumn{2}{c|}{
                \raisebox{-0.15\height}{\includegraphics[width=0.016\linewidth]{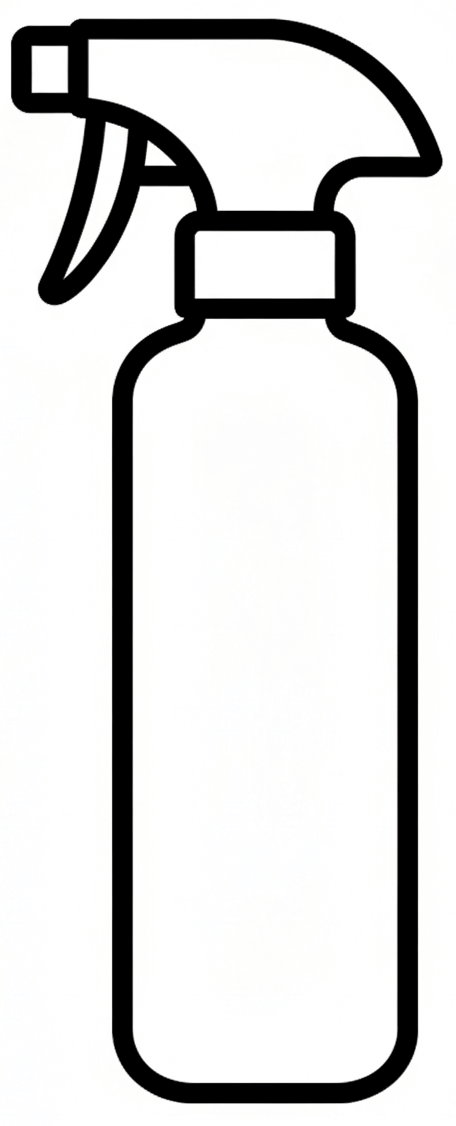}}
                \raisebox{-0.15\height}{\includegraphics[width=0.023\linewidth]{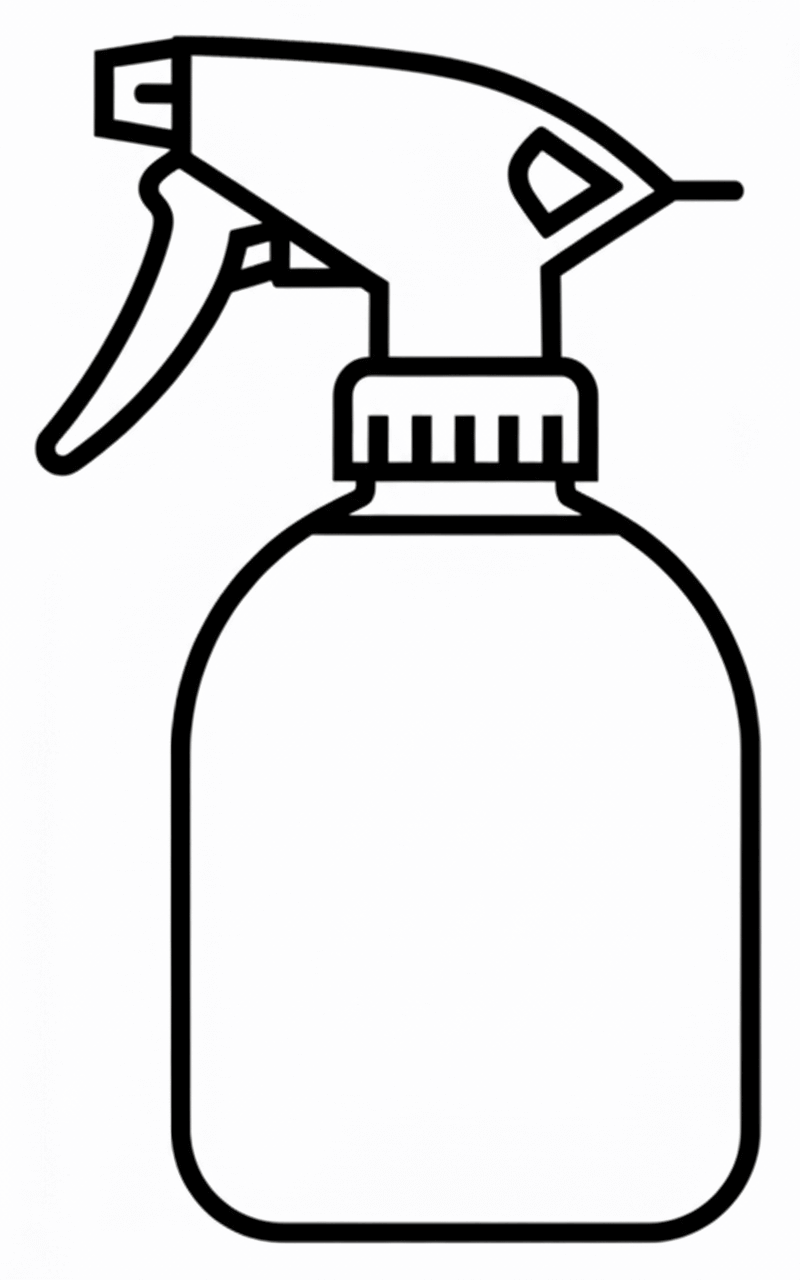}}
                \raisebox{-0.15\height}{\includegraphics[width=0.020\linewidth]{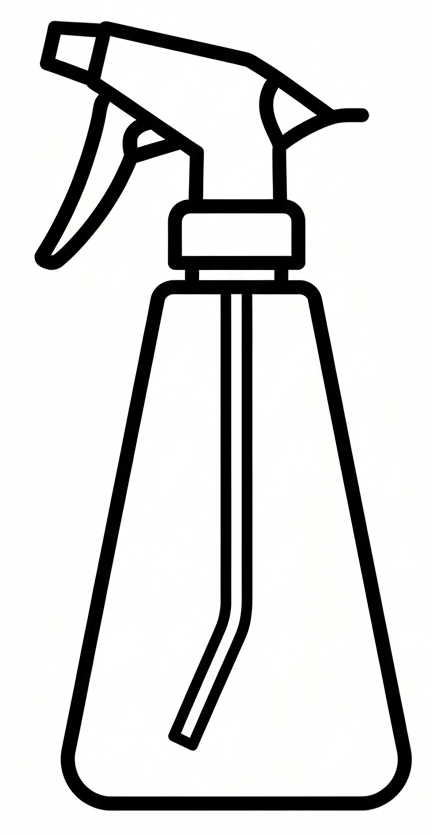}}
            }
          & \multicolumn{2}{c|}{
                \raisebox{-0.15\height}{\includegraphics[width=0.016\linewidth]{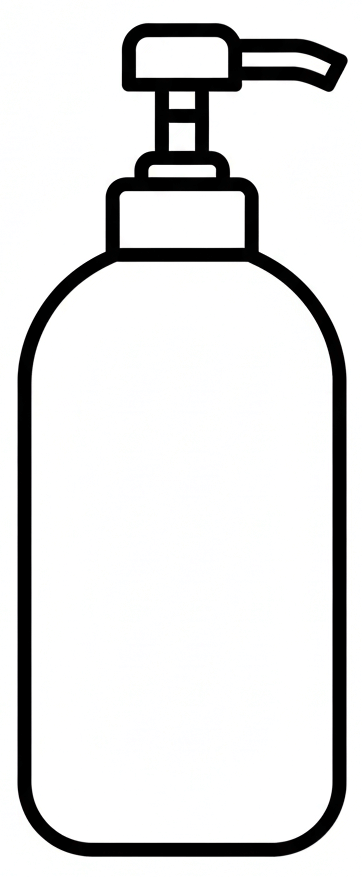}}
                \hspace{2pt}%
                \raisebox{-0.15\height}{\includegraphics[width=0.037\linewidth]{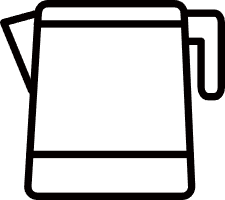}}
                \hspace{2pt}%
                \raisebox{-0.15\height}{\includegraphics[width=0.017\linewidth]{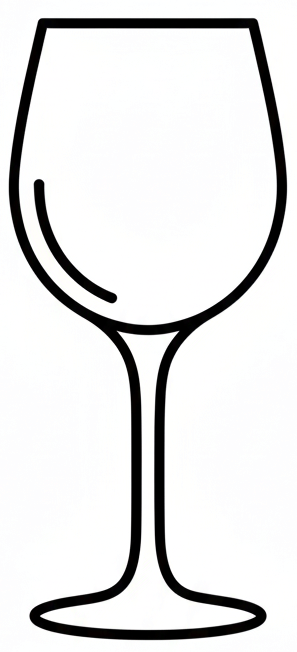}}           
            }
          & \multicolumn{2}{c|}{
                \raisebox{-0.15\height}{\includegraphics[width=0.02\linewidth]{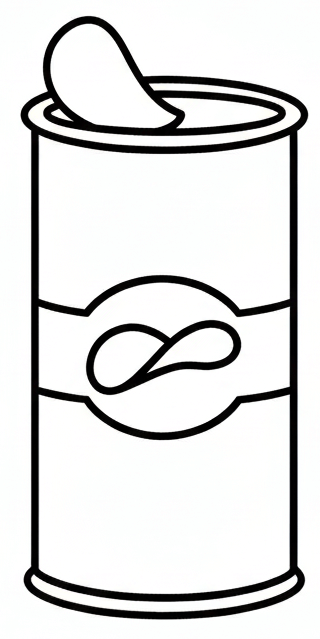}}
                \raisebox{-0.15\height}{\includegraphics[width=0.036\linewidth]{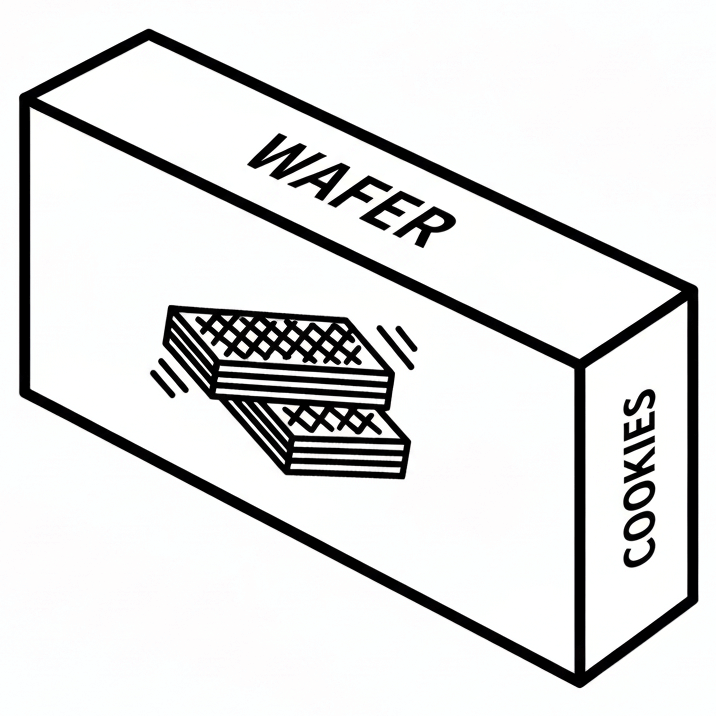}}
                \raisebox{-0.15\height}{\includegraphics[width=0.021\linewidth]{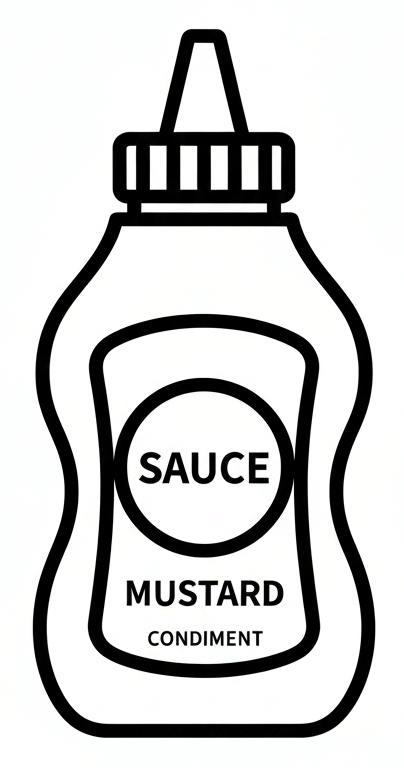}}
            }
          & \multicolumn{2}{c|}{
                \raisebox{-0.15\height}{\includegraphics[width=0.017\linewidth]{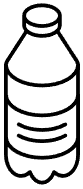}}
                \hspace{2pt}%
                \raisebox{-0.15\height}{\includegraphics[width=0.015\linewidth]{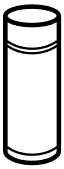}}
                \hspace{2pt}%
                \raisebox{-0.15\height}{\includegraphics[width=0.013\linewidth]{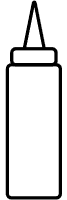}}
            }
          & \multicolumn{2}{c|}{
                \raisebox{-0.15\height}{\includegraphics[width=0.026\linewidth]{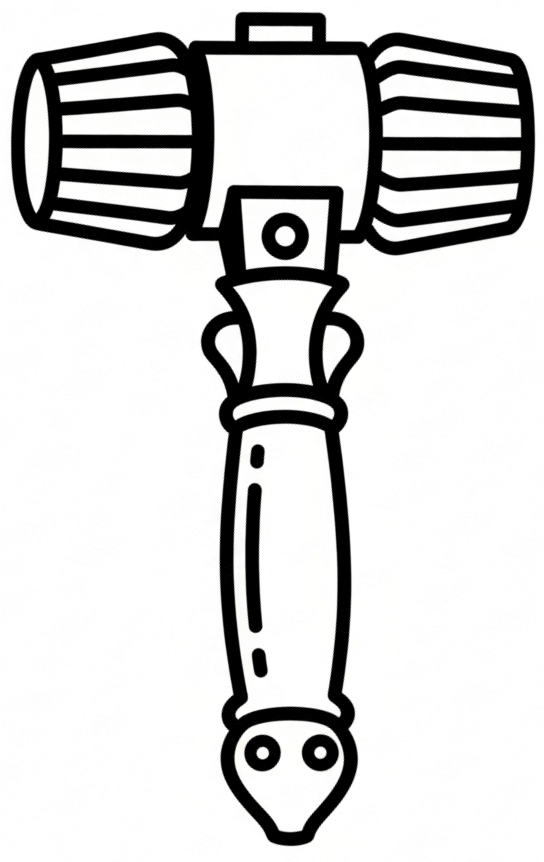}}
                \raisebox{-0.15\height}{\includegraphics[width=0.027\linewidth]{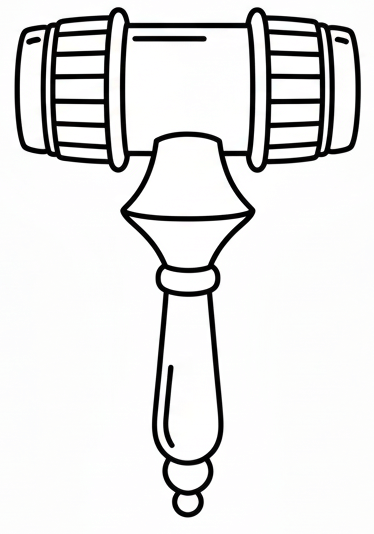}}
                \raisebox{-0.15\height}{\includegraphics[width=0.021\linewidth]{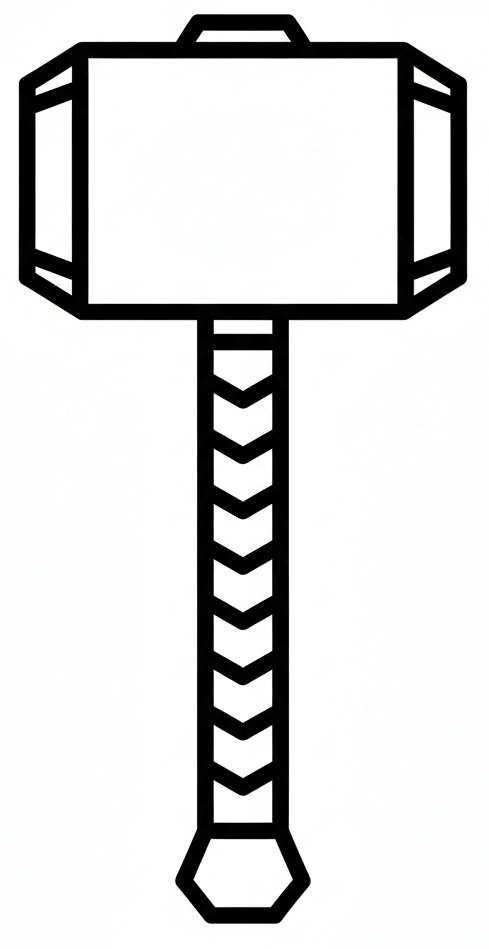}}
            }
          & \multicolumn{2}{c|}{
                \raisebox{-0.15\height}{\includegraphics[width=0.025\linewidth]{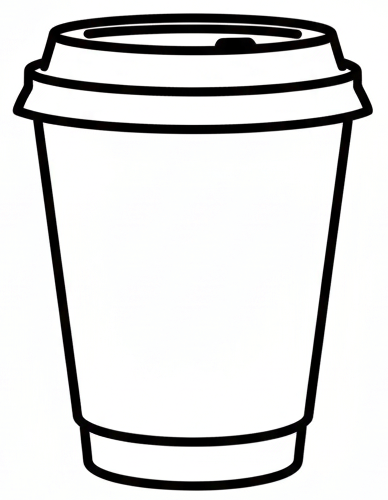}}
                \raisebox{-0.15\height}{\includegraphics[width=0.015\linewidth]{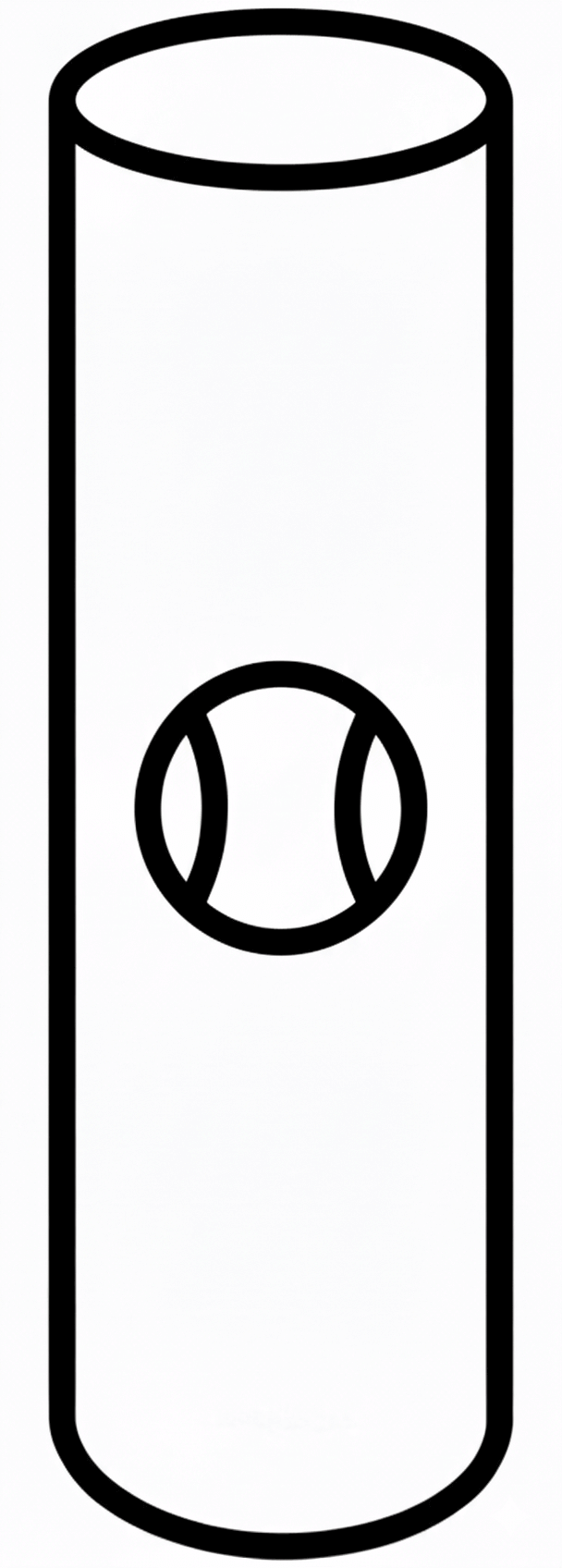}}
                \raisebox{-0.15\height}{\includegraphics[width=0.019\linewidth]{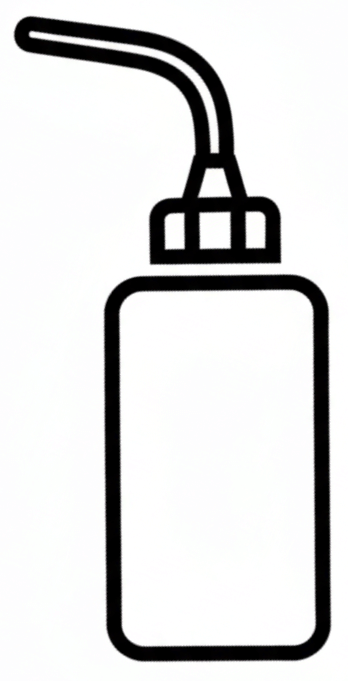}}
            } 
          & \multicolumn{2}{c|}{
                \raisebox{-0.15\height}{\includegraphics[width=0.033\linewidth]{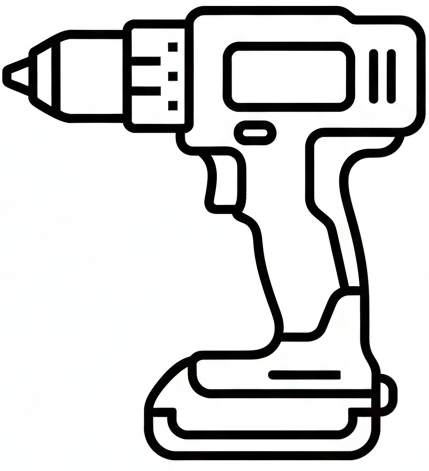}}
                \raisebox{-0.15\height}{\includegraphics[width=0.031\linewidth]{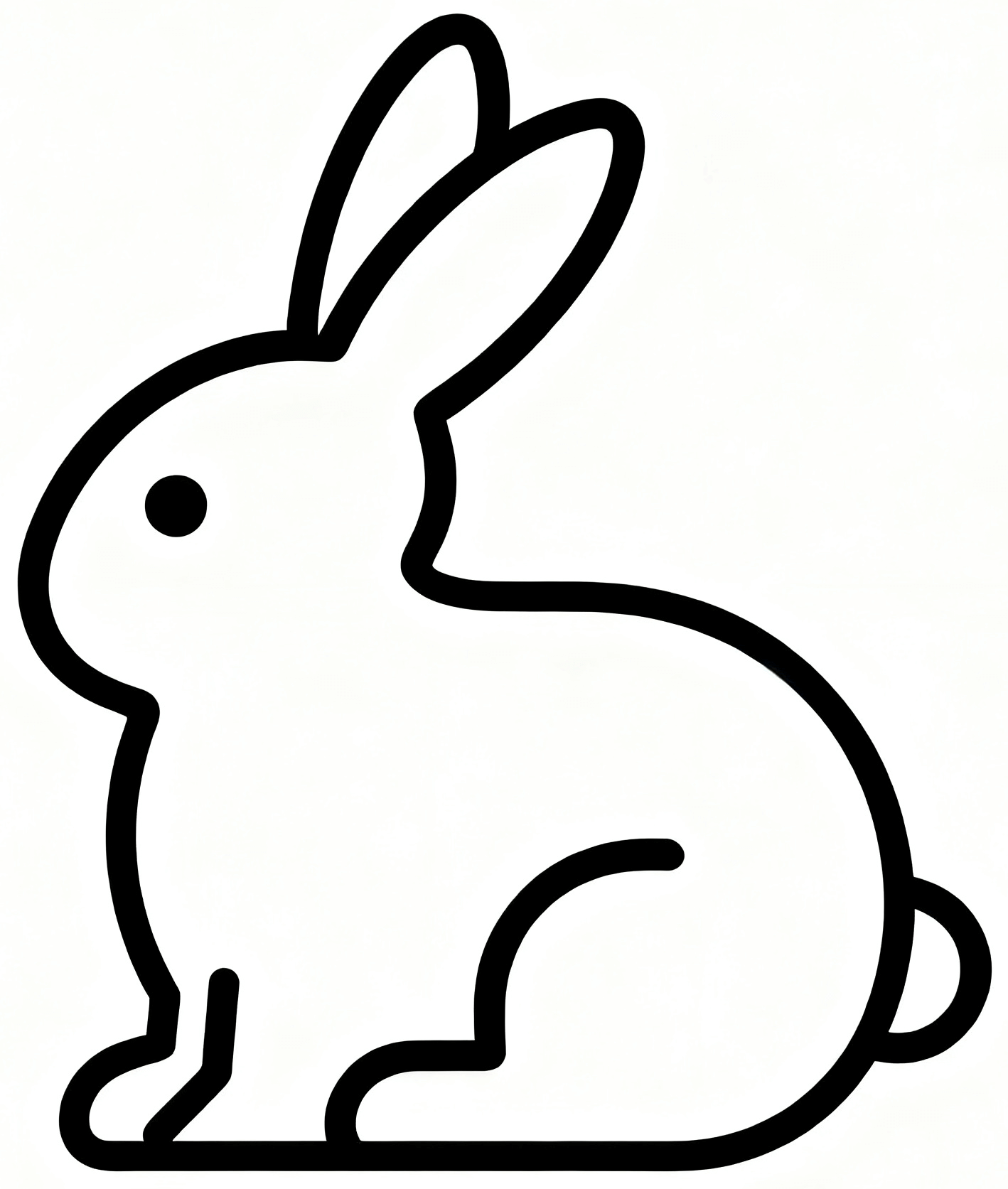}}
                \raisebox{-0.15\height}{\includegraphics[width=0.031\linewidth]{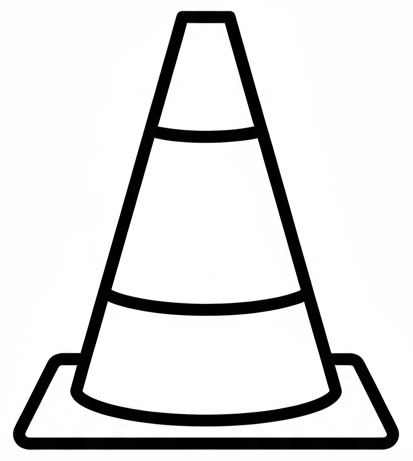}}
            }
            \\ 
\cmidrule(lr){2-3} \cmidrule(lr){4-5} \cmidrule(lr){6-7} \cmidrule(lr){8-9} \cmidrule(lr){10-11} \cmidrule(lr){12-13} \cmidrule(lr){14-15} \cmidrule(lr){16-17} &
            \multicolumn{1}{c}{Suc.} & \multicolumn{1}{c|}{Inten.} &
            \multicolumn{1}{c}{Suc.} & \multicolumn{1}{c|}{Inten.} &
            \multicolumn{1}{c}{Suc.} & \multicolumn{1}{c|}{Inten.} &
            \multicolumn{1}{c}{Suc.} & \multicolumn{1}{c|}{Inten.} &
            \multicolumn{1}{c}{Suc.} & \multicolumn{1}{c|}{Inten.} &
            \multicolumn{1}{c}{Suc.} & \multicolumn{1}{c}{Inten.} & 
            \multicolumn{1}{c}{Suc.} & \multicolumn{1}{c}{Inten.} &
            \multicolumn{1}{c}{Suc.} & \multicolumn{1}{c}{Inten.} \\ 
\midrule
AffordDex   & 20.0  & 0.80  & 87.5  & 3.75  & 25.0  & 1.00  & 25.0  & 0.71  & 25.0  & 1.88  & 33.3  & 1.50  & 55.6  & 2.44  & 37.5 & 1.63  \\ 
Ours        &  \textbf{60.0} & \textbf{3.40}  & \textbf{94.4} & \textbf{4.60}  & \textbf{80.0} & \textbf{4.20}  & \textbf{58.3}  & \textbf{2.50}  & \textbf{73.3} & \textbf{3.2}  & \textbf{77.8}  & \textbf{2.67}  & \textbf{91.7}  & \textbf{4.25}  & \textbf{100.0}  & \textbf{4.50}  \\ 
\bottomrule
\end{tabular}
\caption{Grasping performance on different object categories in real-world experiments. Each category reports success rate (Suc.) and intention-consistency score (Inten.). The AffordDex \cite{wei2025afforddexgrasp} is trained on the dataset \cite{wei2025afforddexgrasp} following its paper.}
\label{tab:objects_results}
\end{table*}

\begin{table}[]
\begin{tabular}{l|cc|cc|cc}
\toprule
          & \multicolumn{2}{c|}{Seen} &
            \multicolumn{2}{c|}{Similar} &
            \multicolumn{2}{c}{Novel} \\ \cline{2-7} 
          & \multicolumn{1}{c}{Stab.} & \multicolumn{1}{c|}{Inten.} &
            \multicolumn{1}{c}{Stab.} & \multicolumn{1}{c|}{Inten.} &
            \multicolumn{1}{c}{Stab.} & \multicolumn{1}{c}{Inten.} \\ \midrule
SceneDiffuser \cite{huang2023scenediffuser} & 3.59 & 4.25 & 2.55 & 2.05 & 2.09 & 1.95 \\
DexGYS \cite{wei2024graspasyousay}        & 3.95 & 4.67 & 2.95 & 2.22 & 2.83 & 1.88 \\
AffordDex \cite{wei2025afforddexgrasp}     & 4.01 & \textbf{4.85} & 3.58 & 3.08 & 3.25 & 2.12 \\ \midrule
Ours                                      & \textbf{4.29} & 3.82 & \textbf{3.96} & \textbf{3.55} & \textbf{4.20} & \textbf{3.88} \\ \bottomrule
\end{tabular}
\caption{The results on the simulation dataset. “Seen,” “Similar,” and “Novel” denote categories that are seen during training, unseen but similar to the training categories, and unseen novel objects, respectively.}
\label{tab:simulation_results}
\end{table}

\begin{table}[]
\centering
\begin{tabular}{l|ll}
\hline
                 & Stab. & Inten. \\ \midrule
Qwen-Image \cite{wu2025qwen-image}      &3.10      &2.51
           \\
GPT-Image \cite{achiam2023gpt}             & 4.84     &4.25
           \\ \midrule
Wan2.2 \cite{wan2025wan}            & 2.52   & 1.62          \\
Wan2.2* \cite{wan2025wan}            & 3.49     &  2.62         \\
Kling-video \cite{kling2024}     & 4.83     &  4.07         \\ \bottomrule
\end{tabular}
\caption{Generation quality across different foundation generative models.}
\vspace{-1ex}
\label{tab:llm_results}
\end{table}

Table~\ref{tab:objects_results} presents the results in real-world experiments, OmniDexGrasp outperforms prior methods in both grasp stability and intention consistency, owing to the generalization capabilities of foundation models as well as our proposed transfer and force-aware grasping strategies. Table~\ref{tab:simulation_results} presents comparisons with additional baseline methods in simulation, while prior methods achieve high performance on seen categories, their performance drops sharply on similar and novel categories, with average scores falling from 4.59 (seen categories) to 2.45 (similar) and 1.98 (novel) respectively, indicating limited generalization ability. In contrast, OmniDexGrasp maintains robust performance across all splits by leveraging foundation models to learn generalizable grasping rules rather than dataset-specific patterns, enabling effective adaptation to unseen scenarios.

\subsection{Can different foundation generative models meet the requirements for dexterous grasping?}
\label{sec: different foundation}
\begin{figure}[t]
    \centering
    \includegraphics[width=\linewidth]{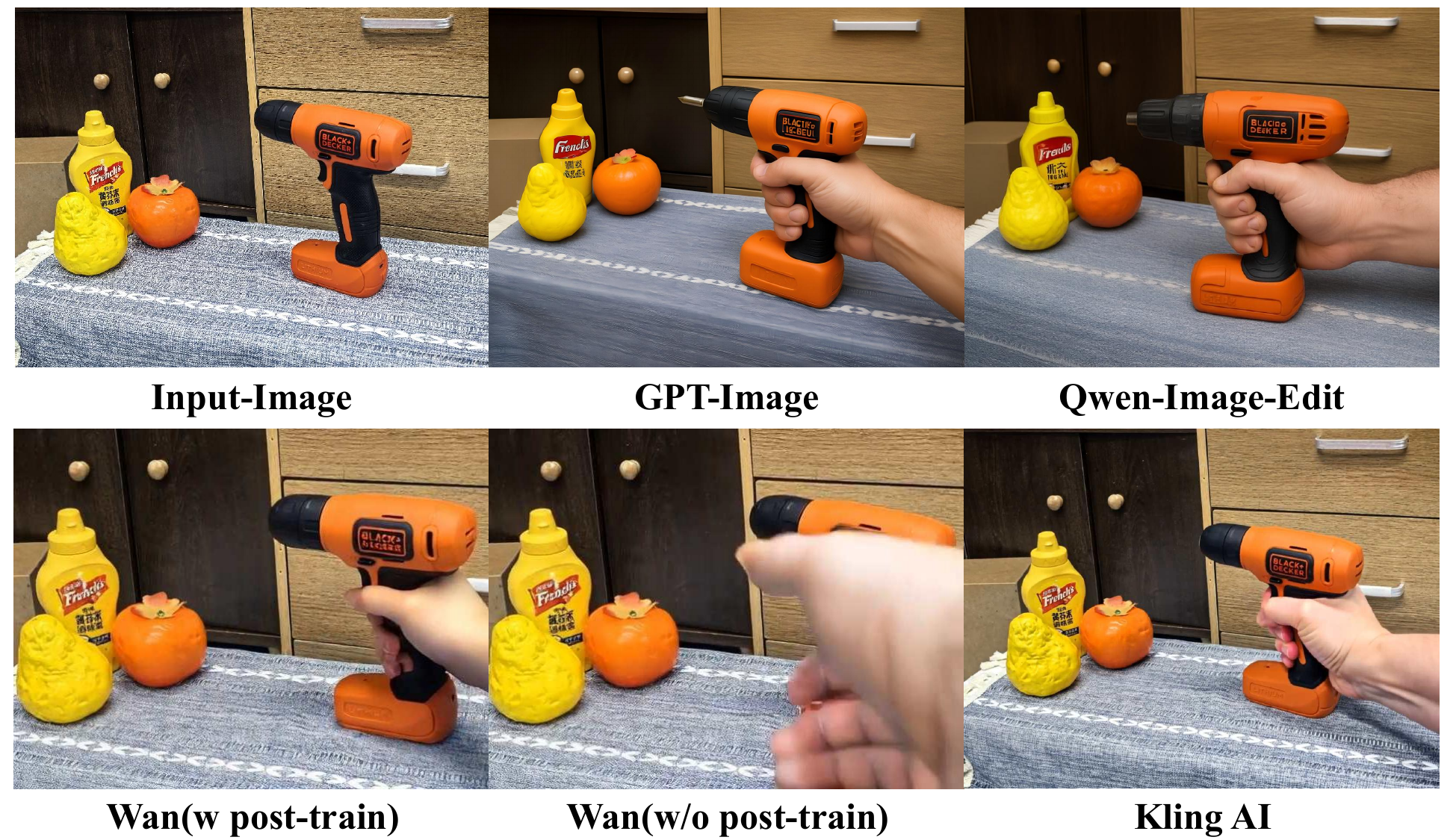}
    \caption{
    The visualization of images generated from different foundation generative models.
    }
    \label{fig:diff_model}
\vspace{-1em}
\end{figure}

We analyze the performance of various foundation generative models to evaluate their suitability for dexterous grasping (Table~\ref{tab:llm_results}, Figure~\ref{fig:diff_model}). Specifically, we provide 20 scene images to different foundation generative models and assess their outputs using GPT-4o. Overall, foundation models demonstrate promising capabilities. We find that closed-source models, such as GPT-Image and Kling-Video, generally outperform open-source models like Owen-Image and Wan2.2. However, we find that the performance of open-source models can improve substantially after fine-tuning on task-specific datasets. For example, Wan2.2* denotes the weights fine-tuned on the EgoDex hand–object interaction dataset \cite{hoque2025egodex}, while Wan2.2 is the original pretrained model, indicating considerable potential for enhancement.

Furthermore, we find that image-based foundation models exhibit stronger instruction-following abilities, reflected in the higher intention-consistency scores of. Conversely, video-based models tend to better in maintaining scene and object pose consistency, which is particularly important for dexterous grasping tasks.

\subsection{Extension to manipulation tasks and cross embodiment}
We conduct qualitative experiments to evaluate the applicability of our approach for achieving cross-embodiment on different dexterous hands and extending it to manipulation tasks beyond dexterous grasping. Visualizations of these experiments are provided in Figure~\ref{fig:setting} and the supplementary video. Benefiting from our human-image-to-robot-action strategy, the framework can be easily adapted to different dexterous hands. Furthermore, by leveraging video generation models \cite{kling2024} and LLM-driven keypoint prediction models \cite{huang2024rekep}, our approach can accomplish several dexterous manipulation tasks. Since grasping forms the fundamental basis of manipulation, we anticipate that our method can be further extended to more complex manipulation scenarios. More visualizations are provided in the supplementary video.

\section{CONCLUSIONS}
We believe that achieving generalizable dexterous grasping with omni-capabilities—across diverse user prompts, dexterous embodiments, and task scenarios—is a central challenge in robotics. To address this, we propose the OmniDexGrasp framework, which leverages foundation models to acquire generalizable understanding ability of scenes and user intentions, and generates human grasp images to guide dexterous grasping. We introduce a human-image-to-robot-action transfer strategy that converts these grasp images into feasible robot actions through hand-object reconstruction and dexterous retargeting. In addition, we design a force-aware adaptive grasping mechanism to ensure stable and safe execution with force feedback. Extensive experiments in both simulation and real-world settings demonstrate the effectiveness of our approach: OmniDexGrasp achieves high success rates and intention-consistency scores across six diverse dexterous grasping tasks, and outperforms state-of-the-art methods on unseen object categories. Overall, our framework highlights the potential of foundation-model-driven approaches for generalizable embodied intelligence.

\begingroup
\interlinepenalty=10000
\bibliographystyle{IEEEtran}
\bibliography{refs}
\endgroup

\end{document}